# Leveraging Large Language Models for Automated Causal Loop Diagram Generation: Enhancing System Dynamics Modeling through Curated Prompting Techniques

Short title: LLM-CLD Generation: Enhancing SD Modeling with Curated Prompting

Ning-Yuan Georgia Liu[1#] and David R. Keith[2,3]

[1] MGH Institute for Technology Assessment, Harvard Medical School, Boston, MA, USA

[2] Melbourne Business School, University of Melbourne, Melbourne, Australia
[3] Sloan School of Management, Massachusetts Institute of Technology, Cambridge, MA, USA

[#] Corresponding Author – ningyuan@vt.edu, gliu26@mgh.harvard.edu

## Abstract

Transforming a dynamic hypothesis into a causal loop diagram (CLD) is crucial for System Dynamics Modelling. Extracting key variables and causal relationships from text to build a CLD is often challenging and time-consuming for novice modelers, limiting SD tool adoption. This paper introduces and tests a method for automating the translation of dynamic hypotheses into CLDs using large language models (LLMs) with curated prompting techniques. We first describe how LLMs work and how they can make the inferences needed to build CLDs using a standard digraph structure. Next, we develop a set of simple dynamic hypotheses and corresponding CLDs from leading SD textbooks. We then compare the four different combinations of prompting techniques, evaluating their performance against CLDs labeled by expert modelers. Results show that for simple model structures and using curated prompting techniques, LLMs can generate CLDs of a similar quality to expert-built ones, accelerating CLD creation.

**Introduction**

The ability to understand the behavior of complex socio-technical systems using System Dynamics (SD) is more important than ever as our world becomes ever-more fast-paced and interconnected. Articulating a formal causal structure and analyzing the behavior of that system using simulation allows the analyst to understand the origins of problematic behavior, and design strategy and policy interventions that effectively address the root causes in the presence of multiple and non-linear feedback loops.

Over the past 70 years, there has been substantial progress in computational and analytical tools that have eliminated multiple burdens for simulation. The field of SD has witnessed significant evolution since its inception by Forrester in the 1950s, enabled by advancements such as the introduction of standardized model formulations, the establishment of robust processes for model development and testing, and the availability of increasingly sophisticated computational tools. Whereas SD models were originally encoded on punch cards and executed on mainframe computers, with behavior-over-time graphs drawn by hand, contemporary SD software packages now allow modelers to create interactive and web-based models that run instantly and which can be calibrated to real-world data using sophisticated analytical methods (Rahmandad et al., 2015). However, the integral process of SD model building - such as representing causal relationships between variables in a Causal Loop Diagram (CLD) - have remained a largely manual process (Sterman, 2000).

Recent advancements in Generative AI have prompted researchers to investigate the application of Large Language Models (LLMs) within the realm of System Dynamics (Jalali & Akhavan, 2024.; Naugle et al., 2024; Veldhuis et al., 2024). Akhavan & Jalali (2023) demonstrate the utility of LLMs in conducting simulation research, concluding that these models can significantly shape and refine idea development and expedite research processes. However, they emphasize that LLMs cannot substitute for critical thinking. Ghaffarzadegan et al. (2024) utilize LLMs to simulate human decision-making in social systems, presenting an innovative method for incorporating dynamic human behaviors into simulations. Furthermore, Hosseinichimeh et al. (2024) introduce a bot that employs LLMs to translate text into graphs, achieving sixty percent accuracy in identifying linkages within two diverse datasets. Nevertheless, their study lacks a detailed explanation of the procedure used for prompting the LLM or the impact of prompting approach on model performance.

CLDs play an integral role in SD modeling because the fundamental assumption in the field built on the structure of the system generates patterns of over-time behavior observed, and CLDs serve as a primary way to communicate the structure of the system (Schaffernicht, 2010). In particular, CLDs represent the causal relationships that exist between variables, focusing the reader on the feedback structure and articulating the "dynamic hypothesis" that the modeler believes explains the patterns of behavior observed. However, CLDs also have noted challenges when it comes to modeling: First, determining the appropriate level of granularity for model boundary objects can be problematic. This challenge is characterized by a lack of precision that may cause modelers to over-aggregate, thereby omitting relevant variables. Second, there is a potential for mislabeling loop polarities, as the dynamics of accumulation are often not fully captured. Third, inferences gained from CLDs will remain speculative without the development of a full simulation model, increasing the risk that inferences are not accurate (Deutsch et al., 2024; Lane, 2008). With these limitations, novice modelers often find the development of high-quality CLDs to be challenging and time-consuming. The availability of better tools to support CLD development could meaningfully accelerate the model building process.

In this paper, we introduce a method for automating the translation of dynamic hypotheses (text) into CLDs using large language models (LLMs) with curated prompting techniques (Chowdhery et al., 2022), leveraging the tremendous recent advancements in Generative Artificial Intelligence (GAI). Significant breakthroughs in Artificial Intelligence and Natural Language Processing have led to the development of LLMs, which demonstrate emergent behaviors such as code generation, story creation, and certain reasoning tasks. Notably, LLMs show promise in performing causal reasoning, which is essential for translating dynamic hypotheses into CLDs in SD modeling. This capability primarily stems from their ability to discern causal relationships from both implicit and explicit cues embedded within texts provided to the model (Hobbhahn et al., 2022; Kosoy et al., 2022; Willig et al., 2022).

Moreover, a recent paradigm shift in the GAI field towards prompt engineering, also referred to as "prompting" - writing textual instructions to guide the model to perform new tasks with minimal data – has further enhanced the utility of LLMs. Few-shot prompting, where the model processes a handful of examples to achieve significant performance, proves advantageous with sparse data (Liu et al., 2021). This method allows LLMs to effectively learn and generate CLDs based on previously unseen dynamic hypotheses, marking a substantial reduction in data dependency and reducing the time to translate from text to CLDs.

We begin by providing an overview of causal reasoning using LLMs, explaining the relevance of these capabilities to translating dynamic hypotheses into CLDs. Next, we describe the collection of a dataset containing high-quality pairs of Dynamic Hypotheses (DH) and CLDs published by leading SD modelers that we use for the purpose of model prompting and testing. We then conduct experiments by running the LLMs on the datasets using four different combinations of prompting techniques, evaluating the LLM-generated CLD against the CLD developed by modelers (i.e., ground truth). We conclude by reflecting on model performance and discussing the potential of this approach to automate more challenging tasks, including the development of more complex CLDs and executable simulation models.

The research we present in this paper is exploratory, intended to introduce the use of LLMs and demonstrate the potential to leverage LLMs as part of the SD toolkit. First, our paper points to new directions for the SD methodology, demonstrating how the process of translating text to a CLD may be automated to accelerate model development and aid novice modelers. Additionally, we provide a demonstration tool – the automatic CLD translator, which is accessible through via the following website: https://cldmaker.azurewebsites.net/. While we focus on simpler CLDs here, the approach we use has the future potential to automate the development of more complex CLDs, and eventually fully executable simulation models. Furthermore, our paper demonstrates how the ability to perform causal reasoning using LLMs has important real-world application in the analysis of complex economic, social and environmental systems.

## Experiment Setup: Generating CLDs from DHs

In this section, we describe the experiment setup, including the LLM model selection, data collection method, and the four combinations of prompting techniques for our LLM model setup.

**Model Selection**

A variety of LLMs exist that differ in the input of textual data on the training dataset, the model size, the hardware used for training, and the training duration (Chowdhery et al., 2022; Zhao et al., 2023). Due to its state-of-the-art reasoning capabilities, this paper uses OpenAI's text-davinci-003 (GPT-3.5) as our backbone LLM model (OpenAI, 2023). All generations are performed using greedy decoding, meaning we take the next-generation sequence's most probable result and disregard all other possibilities (Brown et al., 2020; Ouyang et al., 2022).

**Dataset**

Since no pre-existing dataset of DH-CLD pairs exists appropriate for this task, we developed a dataset containing pairs of high-quality DH-CLD pairs published by leading SD modelers. The dataset (N = 44) contains CLDs ranging from single reinforcing or balancing loops to intermediate complexity CLDs that have 2-4 feedback loops. To build these datasets we used a convenience sampling approach, pulling DH-CLD pairs published in leading SD publications including Business Dynamics (Sterman, 2000), The Systems Thinking Playbook (Sweeney & Meadows, 2010), Modeling the Environment: An Introduction to System Dynamics Models of Environmental Systems (Ford, 1999) and Thinking in Systems (Meadows, 2009) - wherever both a CLD and a clear verbal description of the dynamic hypothesis were present. For each pair, we hand-coded the CLD as a text-based DOT format, a graph description language that captures both causal links and link polarities, which can then be visualized in a Digraph format using the Graphviz package (Ellson et al., 2002). In graph theory, a Digraph, or directed graph, consists of a set of nodes (i.e., variable names) connected by directed edges (i.e., casual links). Moreover, the digraph format also captures the structure and the polarities of a CLD by assigning arrowhead vee **(->)** to denote a positive relationship and arrowhead tee **(-|)** to denote a negative relationship. This structure allows for a representation of both the direction and polarity of the causal relationships in a CLD. A detailed illustration of the Digraph format is provided in Table 1.

Table 1 shows an example of a DH-CLD pair. The first column describes a reinforcing feedback loop where the size of a rabbit population depends on birth rate (Meadows, 2009). The second column shows the actual CLD presented by Meadows. The third column shows our manual transposing of Meadows' CLD into Digraph format generated by the LLM. In the string text Digraph format, each line represents the relationship between two variables, noted in "". A positive polarity between two variables is represented as arrowhead = vee **(->),** while a negative polarity is represented as arrowhead = tee **(-|).** (Note: Meadows' CLD contains a stock and flow, which we don't capture, and lacks polarities on arrows, which we have added into the Digraph for completeness). Lastly, the Digraph format is shown in the fourth column.

*[Insert Table 1. here]*

**LLM Prompting Techniques for CLD Generation**

In our paper, we explore how LLMs can effectively generate human-like CLDs using a variety of curated prompting techniques. One such technique is few-shot prompting, where the model is provided with a small number of input-output examples to guide it towards the desired outputs (Brown et al., 2020). Few-shot prompting has demonstrated significantly better performance compared to zero-shot learning, where no prior examples are given (Brown et al., 2020). The main advantage of few-shot prompting is the reduced need for task-specific data. In our context, where there is a lack of standard data labeling for dynamic hypotheses and CLDs, few-shot prompts allow us to test the feasibility of using LLMs to generate CLDs with minimal data.

To demonstrate the impact of different prompting techniques on model performance, we tested four distinct approaches. The first approach, Baseline, used zero-shot learning, where the LLM received no prior examples. This simulates the typical use of ChatGPT, where the user directly instructs the model to generate a CLD based on the provided dynamic hypothesis without prior examples. This approach tests the LLM's inherent capability to interpret and execute tasks based on its pre-trained knowledge.

Our second approach, Minimal Context, involves few-shot prompting without curated prompts. Here, the LLM receives only the dynamic hypothesis as input, with the expectation of generating the Digraph string format graph. This approach assesses the model's ability to generate structured outputs based solely on minimal context.

In our third approach, Guided Prompts, we enhance the few-shot prompting approach by incorporating curated prompts, that is, specific instructions aimed at guiding the model's response. This involved providing instructions alongside the dynamic hypothesis (illustrated in Guided Prompt), with the output being a structured, labeled graph. Guided Prompt instructs as follows:

> *First, Render a list of variable names from the text given. The variable names should be nouns or nouns phrases. The variable names should have a sense of directionality. Chose names for which the the meaning of an increase or decrease is clear. Second, Render a dot format based on the variable names. A positive relationship is indicated by an arrow from the first variable to the second variable with the sign [vee]. A negative relationship is indicated by an arrow from the first variable to the second variable with the sign [tee].*

This setup is designed to direct the LLM's processing towards a more targeted outcome, mimicking a controlled experimental condition.

In our final approach, Two-Stage, we adopt a Two-stage few-shot prompting strategy. Initially, the LLM is tasked with identifying the relevant variables from the dynamic hypothesis based on the explicit instructions:

> *Render a list of variable names from the text given. Following the rules below = 1. The variable names should be nouns or nouns phrases. 2. The variable names should have a sense of directionality.*

Following this, using the identified variables and the initial dynamic hypothesis, the LLM is prompted to construct the CLD with the second curated prompt:

> *The variables' names will be rendered in DOT format. The steps are as follows: Step 1: Identify the cause-effect relationship between variable names given the dynamic hypothesis. Step 2: [arrowhead=vee] indicates a positive relationship. A negative relationship is indicated by [arrowhead=tee]. Step 3: Create a DOT format based on the cause-effect relationship.*

This two-stage process mimics the thought process of a human system dynamics modeler, focusing first on variable identification and subsequently on mapping out causal relationships. This method is similar to the system dynamics modeling approach outlined by (Sterman, 2000, p152). The sequential flow diagram of Two-stage Approach model setup is depicted in Fig 1.

Through these varied approaches, our study illustrates the adaptability and potential of LLMs in synthesizing dynamic hypotheses, such as text depicting causal relationships to CLDs, which are crucial for system dynamics modeling.

*[Insert Figure 1. Here]*

**Fig 1: Flow Diagram of the Model Setup for Two-stage Approach**

## Results

This section presents the results derived from the four distinct approaches described in the experiment setup using three examples. By employing these different prompting techniques, we compare the DH to the labeled examples (i.e., the ground truth data), which are the CLDs developed by human experts. We present illustrations that demonstrate the behavior of LLMs under each prompting condition. Additionally, these examples highlight how progressively sophisticated prompting techniques enhance the model's performance.

**Example 1: Single reinforcing loop - Smoking cigarettes**

To begin, we use a single reinforcing loop example focused on cigarette addiction (Meadows, 2009). The DH is as follows: *"The more my uncle smokes, the more addicted he becomes to the nicotine in his cigarettes. After smoking a few cigarettes a long time ago, my uncle began to develop a need for cigarettes. The need caused him to smoke even more, which produced an even stronger need to smoke. The reinforcing behavior in the addiction process is characteristic of positive feedback."*

The outcome of the four approaches is shown in Fig 2 (for Approach 1) and Table 2 (for Approaches 2, 3, 4). The ground truth CLD, based on (Meadows, 2009), is displayed in the first column of Table 2. Using a no-prompts prior (Approach 1), we see that the LLM can identify and articulate causal relationships between variables. For instance, it understands the fundamentals of what a CLD is and correctly identifies and describes a positive feedback loop between "*smoking*" and "*addiction*", including the direction of the cause-effect relationship between these variables. This demonstrates the LLM's capability to discern causal connections, though it does not yet generate a graphical representation. This insight is a promising indicator of the LLM's potential to understand and represent dynamic complexity.

*[Insert Figure 2. Here]*

**Fig 2. Generated Outputs for Cigarettes Addiction from Approach 1**

The LLM generated a graphical representation when we used a few-shot learning prompt (Approach 2). While most variable names were accurately identified, the exogenous variable "*addiction time*" was not accurately recognized. Despite this, the conceptual understanding of the reinforcing loop between "*smoking*" and "*addiction*" was accurate. However, the model generated by the LLM included a negative relationship between the "need for cigarettes" and "smoking," which is contrary to the statements in the DH.

Next, implementing the Guided Prompts Approach, utilizing curated prompts in addition to the dynamic hypothesis and the expected output (ground truth data), resulted in correct identification of the "need for cigarettes" variable. However, this model included "reinforcing behavior" as a variable, indicating that the LLM can recognize the reinforcing behavior of addiction in cigarettes, but treated it as a variable name rather than a description of the loop. Approach 4 demonstrated a marked improvement in the model's performance. Here, the model accurately identified all relevant variables and correctly recognized the reinforcing dynamics. Notably, the LLM successfully detected the exogenous variable "addiction time". However, it struggled to accurately determine the relationship between "addiction time" and "need for cigarettes" This limitation likely stems from the model's difficulty in processing exogenous variables that are less central to the primary DH, which is not explicitly documented in the text. This suggests an area for further prompt refinement in how the LLM handles variables that are not directly emphasized within the DH, and also underscores the need for clear articulation of causal relationship when writing the DH. Future research may explore a new way of writing DH, since exogenous variables can also have a significant impact on the behavior of the system.

*[Insert Table 2. here]*

**Example 2:  Two Balancing Loops – New Car Inventory**

In our second example, we developed into a more complex case involving two balancing loops related to new car inventory based on (Ford, 1999). The dynamic hypothesis is as follows: "*Car production builds the inventory of cars at the dealer. A higher inventory leads to a lower market price, and lower market prices cause less car production in the future. If the price were to increase, the retail sale of cars would tend to fall. Retails sales drain the inventory of cars held in stock at the dealership. And a decline in the inventory will cause the dealers to raise their prices in the future.*"

The outcomes of the four approaches for this example are shown in Fig 3 (for Approach 1) and Table 3 (for Approach 2, 3, and 4), with the ground truth represented in the first column of Table 3. In Approach 1, the model captured the causal relationships between variables even without specific prompts, though the structure and clarity of the output could be enhanced (Fig 3). While it successfully identified the balancing feedback loop concerning car production, inventory, and market prices, it missed the second balancing loop involving retail sales and market price.

Approach 2, which employed few-shot prompting, only partially recognized the variable names, with some overlap, mislabeling  "*inventory of cars at the dealership*" as simply "*inventory*" and failing to consolidate "market price" and "price" into a single variable. Although there was a significant improvement in terms of capturing the feedback loop, the feedback loops presented in this version contained link inaccuracies.  Approach 3 further improved the output, as the LLM accurately identified two negative feedback loops. However, some causal relationships between variables were still inaccurately depicted.

Lastly, Approach 4 presented a significant improvement in quality again. This latest iteration demonstrated precise identification of variables and more accurate depiction of causal relationships, though it incorrectly identified the negative relationship between "*Market price*" and "*Retail car sales*". This shortcoming likely stems from insufficient specificity in the DH, indicating the need for more detailed inputs and curated prompts to improve model performance in generating CLDs.

In this more complex example containing multiple feedback loops, it becomes evident that a significant enhancement in model performance is achieved through a combination of curated prompts and few-shot prompting techniques (Approach 4). This approach leverages detailed guidance to the LLM and specific examples, thereby generating a more accurate representation of human-like CLD.

*[Insert Figure 3. Here]*

**Fig 3. Generating Output for New Car Inventory from Approach 1**

*[Insert Table 3. Here]*

**Example 3: Two Balancing Loop with Exogenous Variables - Assignment Backlog Dynamics**

In the third example, we explored a much more complex CLD containing two balancing loops with multiple exogenous variables to illustrate the dynamics of assignment backlogs (Sterman, 2000, p. 164). This example introduces a CLD with comprehensive descriptions and expanded DH, enhancing the readers' understanding of the underlying dynamics of the process. The DH we used was: "*The Assignment Backlog is increased* *by the Assignment Rate and decreased by the Completion Rate. Completion Rate is Workweek (hours per week) times Productivity (tasks completed per hour of effort) times the Effort Devoted to Assignments. Effort Devoted to Assignments is the effort put in by the student compared to the effort required to complete the assignment with high quality. If work pressure is high, the student may choose to cut corners, skim some reading, skip classes, or give less complete answers to the questions in assignments. For example, if a student works 50 hours per week and can do one task per hour with high quality but only does half the work each assignment requires for a good job, then the completion rate would be (50)(1)(.5) = 25 task equivalents per week. Work Pressure determines the workweek and effort devoted to assignments. Work pressure depends on the assignment backlog and the Time Remaining to complete the work: The bigger the backlog or the less time remaining, the higher the workweek needs to be to complete the work on time. Time remaining is of course simply the difference between the Due Date and the current Calendar Time. The two most basic options available to a student faced with high work pressure are to first, work longer hours, thus increasing the completion rate and reducing the backlog , or second, work faster by spending less time on each task, speeding the completion rate and reducing the backlog. Both are negative feedbacks whose goal is to reduce work pressure to a tolerable level."*

Fig 4 (for Approach 1) and Table 4 (for Approach 2, 3, and 4) display the outcomes of the four iterations. Similar with the previous examples, the result of Approach 1 successfully captured the two negative feedback loops in the dynamic hypothesis, albeit in textual rather than graphical format. Moreover, it appears that the model is (unsuccessfully) trying to generate links between variables, underscoring one of the emerging capabilities of LLMs—the extraction of causal relationships from text.

[Insert Fig 4. Here]

**Fig 4. Generated Outputs for Assignment Backlog from Approach 1**

In Approach 2, the LLM accurately recognizes variable names, yet fails to identify any feedback loops. Furthermore, the relationship between the "*work completion rate*" and the "*assignment backlog*" is incorrect. Nevertheless, this iteration demonstrates LLM's capability to identify most exogenous variables, even though they are not central to the main feedback dynamics. Introducing a curated prompt in Approach 3 leads to modest enhancements in the performance. Notably, the connection between "*assignment backlog*" and "*work pressure*" is correctly identified, indicating a step towards improved model performance.

Lastly, the two-stage approach in Approach 4 delivers a substantial improvement in output quality. In this iteration, the model not only identified a feedback loop but also achieves precise identification of variables and accurately depicts casual links. This iteration demonstrated the LLM's enhanced ability to model complex diagrams with surprising accuracy when equipped with curated and designed prompting techniques. Moreover, in this iteration, the directionality between the "*work completion rate*" and the "*assignment backlog*" was correctly identified, demonstrating that a higher work completion rate leads to a reduced assignment backlog. Nevertheless, certain relationships remain elusive to the model. For example, the link between "*work pressure*" and the "*effort to develop assignments*" was overlooked, a crucial component of the second balancing loop. This suggests that the two-stage approach employed may assist the LLM in sequentially processing causal relationships, somewhat mimicking the reasoning of a human modeler. Despite these advancements, there is still room for further refinement.

[Insert Table 4 Here]

## Discussion: Challenges and Future Research Opportunities

LLMs have gained considerable attention in recent times given their promise to transform what tasks humans are able to undertake and augment human effort. In this paper we demonstrated that LLMs also have the potential to accelerate our ability to analyze complex systems, automating the development of Causal Loop Diagrams central to the model development process in SD. Our contribution is in three folds: first, our results show that LLMs are able to generate CLDs comparable to those built by expert human modelers for simple feedback structures, particularly when the dynamic hypothesis is articulated clearly and in detail. Second, with curated prompting techniques, LLMs are able to generate CLDs that are more closely to CLDs developed by modelers. Third, the development of this capability in the SD toolkit has potential benefits including accelerating the process of CLD development, overcoming barriers for novice modelers that will aid the growth of the SD field, and driving quality standards in SD modeling.

This capability may be particularly useful for example in applications such as Group Model-Building (GMB) sessions (Hovmand, 2014; Peck, 1998), where a vast amount of interview transcripts need to be translated into mental maps and CLDs. GMB sessions are collaborative workshops where stakeholders and modelers work together to construct models that represent the underlying structure and behavior of complex systems. These sessions often generate extensive qualitative data, including interview transcripts, notes, and discussions, which must be meticulously analyzed and synthesized into coherent models. Automating the translation of these qualitative data into CLDs can significantly enhance the efficiency and effectiveness of GMB sessions. It reduces the time and effort required for manual data analysis and model construction, allowing participants to focus more on validating and enhancing the model rather than on its initial development. Overall, the integration of LLMs into the SD toolkit can revolutionize the way GMB sessions are conducted, making them more efficient, effective, and inclusive.

However, considerable is still needed to refine and test an LLM tool that can be confidently used to develop high-quality CLDs. LLMs are a black box, and while we demonstrate the apparent ability of LLMs to infer causal relationships and apply it to the development of CLDs, we have limited ability to explain why the model generates output in the way it does, or what approaches will elevate performance on translating dynamic hypothesis to CLD tasks. Continued effort is needed to test different prompts and prompting strategies (e.g., Automated Prompt Engineering that uses gradient search to refine prompts) to achieve better performance. Ideally, this would also involve a set of best practices for using LLMs to develop CLDs.

As novel as our results are, they are based on simple CLDs that are unrealistically small for many real-world applications. Our intuition is that using larger datasets containing more complex CLDs will ultimately yield higher levels of performance, but this is yet to be demonstrated. Related, in the collection of examples for the datasets we build in this paper, we observed that complex CLDs frequently lack a succinct articulation of the dynamic hypothesis. Descriptions were often embedded in longer passages of text such as book chapters, newspaper articles and corporate reports. Therefore, best practices of using LLMs in automating the generation of CLDs may include an approach for analysts to more clearly articulate dynamic hypotheses using standardized language that lends itself to translation.

Finally, while we use LLMs to generate CLDs in this paper, the approach we use should in future be able to generate fully executable model code in XMILE or other formats, just as LLMs have been shown capable of generating code in general purpose programming languages. To do so, it would be necessary to provide the LLM additional information not used here, such as reference modes, data sources, and initial conditions for stock variables.

The ultimate strength of System Dynamics lies in the ability to understand complex patters of system behavior over time through the articulation of an explicit causal structure. Rather than using AI to predict the behavior of systems into the future - a zero-th order' approach that fails to make use of information about the underlying structure of the system - we propose the use of AI specifically for the elicitation of system structure. Leveraging the incredible recent developments in LLMs can both help modelers with the time-consuming and intensive process of model development, and increase the accessibility and reach of System Dynamics as a discipline, improving our ability to manage complex systems in business, public policy, and society.

## Acknowledgments

We thank Dr. XXXX and Dr. XXXX for their constructive feedback on earlier versions of this paper.

# References

Akhavan, A., & Jalali, M. S. (2023). Generative AI and simulation modeling: How should you (not) use large language models like ChatGPT. *System Dynamics Review*.

Brown, T. B., Mann, B., Ryder, N., Subbiah, M., Kaplan, J., Dhariwal, P., Neelakantan, A., Shyam, P., Sastry, G., Askell, A., Agarwal, S., Herbert-Voss, A., Krueger, G., Henighan, T., Child, R., Ramesh, A., Ziegler, D. M., Wu, J., Winter, C., & Amodei, D. (2020). *Language Models are Few-Shot Learners*. https://papers.nips.cc/paper_files/paper/2020/file/1457c0d6bfcb4967418bfb8ac142f64a-Paper.pdf

Chowdhery, A., Narang, S., Devlin, J., Bosma, M., Mishra, G., Roberts, A., Barham, P., Chung, H. W., Sutton, C., Gehrmann, S., Schuh, P., Shi, K., Tsvyashchenko, S., Maynez, J., Rao, A., Barnes, P., Tay, Y., Shazeer, N., Prabhakaran, V., & Fiedel, N. (2022). *PaLM: Scaling Language Modeling with Pathways*. http://arxiv.org/abe/2204.02311

Deutsch, A. R., Frerichs, L., Perry, M., & Jalali, M. S. (2024). Participatory modeling for high complexity, MULTI-SYSTEM issues: Challenges and recommendations for balancing qualitative understanding and quantitative questions. *System Dynamics Review*, sdr.1765. https://doi.org/10.1002/sdr.1765

Ellson, J., Gansner, E., Koutsofios, L., North, S. C., & Woodhull, G. (2002). Graphviz—Open Source Graph Drawing Tools. In P. Mutzel, M. Jünger, & S. Leipert (Eds.), *Graph Drawing* (pp. 483–484). Springer. https://doi.org/10.1007/3-540-45848-4_57

Ford, F. A. (1999). *Modeling the environment: An introduction to system dynamics models of environmental systems*. Island press.

Ghaffarzadegan, N., Majumdar, A., Williams, R., & Hosseinichimeh, N. (2024). Generative agent-based modeling: An introduction and tutorial. *System Dynamics Review*, *40*(1), e1761. https://doi.org/10.1002/sdr.1761

Hobbhahn, M., Lieberum, T., & Seiler, D. (2022). *Investigating causal understanding in LLMs*. https://openreview.net/pdf?id=FQ15KxgFRc

Hosseinichimeh, N., Majumdar, A., Williams, R., & Ghaffarzadegan, N. (2024). *From Text to Map: A System Dynamics Bot for Constructing Causal Loop Diagrams* (arXiv:2402.11400). arXiv. http://arxiv.org/abs/2402.11400

Hovmand, P. S. (2014). Group Model Building and Community-Based System Dynamics Process. In P. S. Hovmand (Ed.), *Community Based System Dynamics* (pp. 17–30). Springer. https://doi.org/10.1007/978-1-4614-8763-0_2

Jalali, M. S., & Akhavan, A. (2024). Integrating AI language models in qualitative research: Replicating interview data analysis with ChatGPT. *System Dynamics Review*, *n/a*(n/a). https://doi.org/10.1002/sdr.1772

Kosoy, E., Chan, D. M., Liu, A., Collins, J., Kaufmann, B., Huang, S. H., Hamrick, J. B., Canny, J., Ke, N. R., & Gopnik, A. (2022). *Towards Understanding How Machines Can Learn Causal Overbypotheses*. http://arxiv.org/abs/2206.08353

Lane, D. C. (2008). The emergence and use of diagramming in system dynamics: A critical account. *Systems Research and Behavioral Science*, *25*(1), 3–23.

Liu, P., Yuan, W., Fu, J., Jiang, Z., Hayashi, H., & Neubig, G. (2021). *Pre-train, Prompt, and Predict: A Systematic Survey of Prompting Methods in Natural Language Processing*. http://arxiv.org/abs/2107.13586

Meadows, D. H. (2009). *Thinking in Systems: A Primer*. Earthscan.

Naugle, A., Langarudi, S., & Clancy, T. (2024). What is (quantitative) system dynamics modeling? Defining characteristics and the opportunities they create. *System Dynamics Review*, *40*(2), e1762. https://doi.org/10.1002/sdr.1762

OpenAI. (2023). *OpenAI GPT-3 API [text-davinci-003*. https://platform.openai.com/docs/api-reference

Ouyang, L., Wu, J., Jiang, X., Almeida, D., Wainwright, C. L., Mishkin, P., Zhang, C., Agarwal, S., Slama, K., Ray, A., Schulman, J., Hilton, J., Kelton, F., Miller, L., Simens, M., Askell, A., Welinder, P., Christiano, P., Leike, J., & Lowe, R. (2022). *Training language models to follow instructions with human feedback*. http://arxiv.org/abs/2203.02155

Peck, S. (1998). Group model building: Facilitating team learning using system dynamics. *Journal of the Operational Research Society*, *49*(7), 766–767.

Rahmandad, H., Oliva, R., & Osgood, N. D. (2015). *Analytical Methods for Dynamic Modelers*. MIT Press.

Schaffernicht, M. (2010). Causal loop diagrams between strusture and behaviour A critical analysis of the relationship between pelarity, behaviour and events. *Systems Research and Behavioral Science*, *27*(6), 653–666.

Sterman, J. (2000). *Business Dynamics: Systems Thinking and Modeling for a Complex World*. McGraw-Hill Education.

Sweeney, L. B., & Meadows, D. (2010). *The Systems Thinking Playbook: Exercises to Stretch and Build Learning and Systems Thinking Capabilities*. Chelsea Green Publishing.

Veldhuis, G. A., Blok, D., de Boer, M. H. T., Kalkman, G. J., Bakker, R. M., & van Waas, R. P. M. (2024). From text to model: Leveraging natural language processing for system dynamics model development. *System Dynamics Review*, *n/a*(n/a), e1780. https://doi.org/10.1002/sdr.1780

Willig, M., Zeevic, M., Dhami, D. S., & Kersting, K. (2022). *Can Foundation Models Talk Causality?* http://arxiv.org/abs/2206.10591

Zhao, Z., Song, S., Duah, B., Macbeth, J., Carter, S., Van, M. P., Bravo, N. S., Klenk, M., Sick, K., & Filipowicz, A. L. S. (2023). More human than human: LLM-generated narratives outperform human-LLM interleaved narratives. *Proceedings of the 15th Conference on Creativity and Cognition*, 368–370.

**Table 1. Example DH-CLD pair provided to the LLM**

| Input: Dynamic Hypothesis | Output: Causal Loop Diagram | Output: CLD – Digraph String format | Output: CLD - Digraph |
|---|---|---|---|
| The larger the population, the greater the number of births. increases, the faster the population increases. The more the birth rate increases, the faster the population increases. | 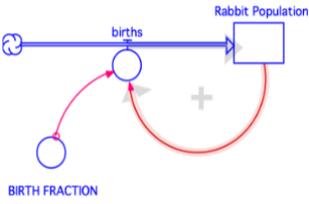 | digraph { "births" -> "rabbit population" [arrowhead = vee] "rabbit population"->"births"[arrowhead = vee] "birth fraction" -> "births"[arrowhead = vee] } | 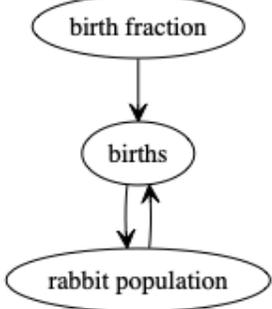 |

**Table 2.  Generated Outputs for Cigarettes Addiction from Approachs 2-4**

| Label graph | Approach 2 | Approach 3 | Approach 4 |
|---|---|---|---|
| 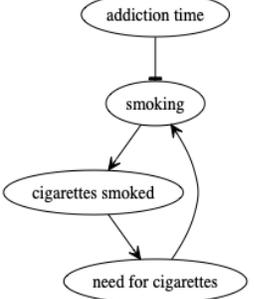 | 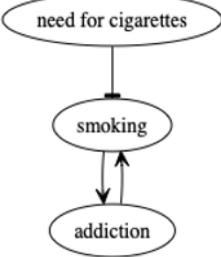 | 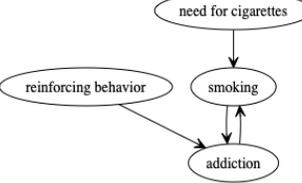 | 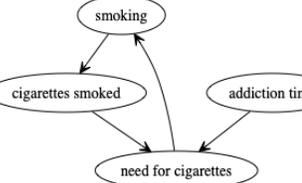 |

## Table 3. Generating Output for New Car Inventory from Approaches 2-4

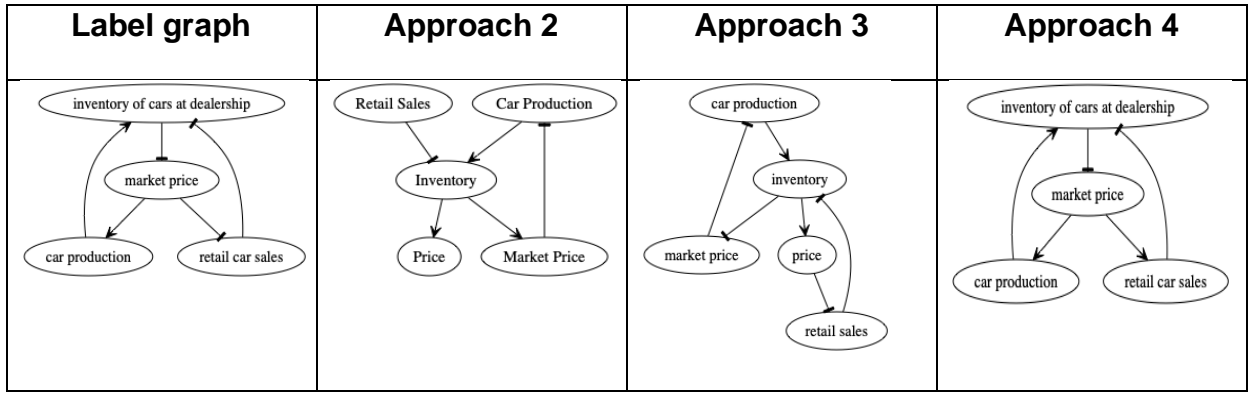

## Table 4.    Generated Outcome of Assignment Backlog

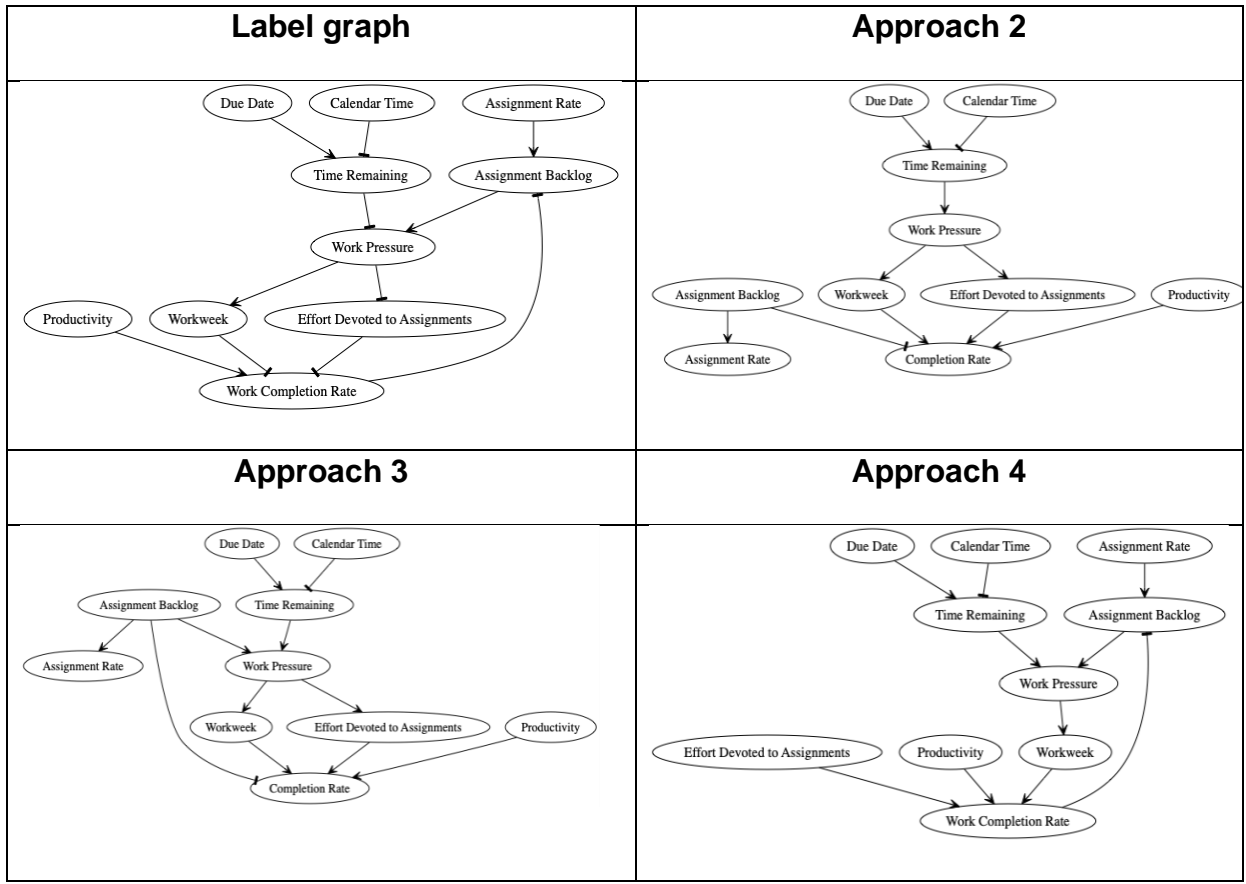